\documentclass[11pt]{article}

\usepackage[preprint]{acl}

\usepackage{times}
\usepackage{latexsym}
\usepackage{todonotes}
\usepackage{multirow} 
\usepackage{arydshln}
\usepackage{booktabs}    
\usepackage{graphicx}    
\usepackage{geometry}

\usepackage[most]{tcolorbox}
\usepackage{xcolor}
\usepackage{listings}
\tcbset{
  promptbox/.style={
    enhanced, breakable, colback=gray!5, colframe=black!70,
    boxrule=0.6pt, arc=2pt, left=6pt, right=6pt, top=4pt, bottom=4pt,
    fonttitle=\bfseries\small, coltitle=white, colbacktitle=black!70,
    attach boxed title to top left={xshift=4pt, yshift=-2pt},
    boxed title style={colback=black!70, sharp corners},
    fontupper=\small\ttfamily,
  }
}

\usepackage{amsmath} 
\usepackage{amssymb}

\usepackage[T1]{fontenc}

\usepackage[utf8]{inputenc}

\usepackage{microtype}

\usepackage{inconsolata}

\usepackage{graphicx}
\usepackage{amsmath}
\title{EAPO: Entropy-Driven Adaptive Positive-Negative Sample Weighting for Policy Optimization in Open-Ended QA}

\author{
Yunsheng Zeng$^{1,}\thanks{\hspace{2pt}Equal contribution. E-mail: zengyunsheng@bupt.edu.cn}$,
Gen Li $^{1,}$\footnotemark[1], 
Yuwei Miao$^{1,}$\footnotemark[1],
Xiandong Li$^{1}$, 
Yujin Wang$^{1}$,
Siyu Chen$^{1}$,\\
\textbf{Luning Wang}$^{1}$,
\textbf{Yunhao Qiao}$^{1}$,
\textbf{Junfeng Wang}$^{1}$,
\textbf{Jianwei Lv}$^{1,}\thanks{\hspace{2pt}Corresponding author. E-mail: \{yuanbo07,lvjianwei\}@baidu.com}$, 
\textbf{Bo Yuan}$^{1,}$\footnotemark[2]  \\
$^{1}$Baidu Inc.
}

\begin{document}
\maketitle
\begin{abstract}
Large Reasoning Models are typically trained via reinforcement learning from verifiable rewards (RLVR). However, existing approaches adopt fixed weights for positive and negative samples, and the conclusions hardly generalize to open-ended question answering (QA).
In this paper, we systematically investigate the roles of positive and negative samples in reinforcement learning for open-ended QA. We propose a reward-mean-based strategy for distinguishing positive from negative samples, and observe that negative samples predominantly govern response diversity and the performance upper bound, whereas positive samples primarily determine response quality and convergence stability.
Building on these observations, we propose \textbf{EAPO}, an \textbf{E}ntropy-driven \textbf{A}daptive \textbf{P}olicy \textbf{O}ptimization method that adaptively computes the weighting coefficients of positive samples based on the ratio of the current policy entropy to the initial entropy. During the entropy-decreasing phase, the weight assigned to positive samples is reduced to preserve exploration, whereas during the entropy-increasing phase it is amplified to reinforce stability, thereby mitigating entropy collapse. Experiments on two publicly available open-ended medical QA datasets demonstrate that EAPO consistently and substantially outperforms fixed-weight baselines in both response diversity and stability.
\end{abstract}

\section{Introduction}
Large Reasoning Models (LRMs) have recently attracted considerable attention owing to their remarkable performance on mathematical\citep{mirzadeh2025gsm,wang2026survey}, coding\citep{chen2025steering,yang2025code}, and scientific reasoning tasks\citep{yan2025position,phan2025humanity}. Such models are typically trained via reinforcement learning from verifiable rewards (RLVR) \citep{guo2025deepseek,team2025kimi,huang2026direction,wu2026quantile,xu2026understanding,ingle2026adaptive,xu2026you}, where multiple long chain-of-thought trajectories are sampled and assigned a binary reward according to final-answer correctness, and the policy is then updated based on reward signal. 

Recent studies\citep{tang2025rethinking,zhu2026surprising} have investigated the impact of positive and negative samples in RLVR. \citet{zhu2026surprising} finds that using only positive samples improves Pass@1 but harms diversity, leading to a decrease in Pass@k for higher k, whereas using only negative samples maintains exploration and yields stronger performance at high k. \citet{tang2025rethinking} further explores adjusting the influence strength of positive and negative samples at different granularities (sample-level and token-level), thereby modulating the training process of RLVR.

However, existing policy optimization methods \citep{zhu2026surprising} typically employ fixed weights for positive and negative samples. When the weight assigned to positive samples is excessively high, the policy is prone to entropy collapse, which leads to a lack of diversity in generated outputs and undermines the model's exploration capability. Conversely, when the weight on negative samples is too large, the exploration space expands considerably, causing the model to exhibit persistent instability throughout training. Although \citet{tang2025rethinking} introduces asymmetric advantage shaping over positive and negative samples at the token level during training, its experimental conclusions are confined to the RLVR setting and thus fail to generalize to open-ended question answering (QA) scenarios.
Owing to the vast solution space and the absence of objective ground-truth answers, the distinct roles played by positive and negative samples in open-ended QA remain insufficiently explored.



In this paper, we systematically investigate the roles of positive and negative samples in reinforcement learning for open-ended QA. Unlike RLVR, open-ended QA lacks fixed ground-truth answers against which positive and negative samples can be readily distinguished. To address this issue, we devise a reward-mean-based strategy for selecting positive and negative samples. We observe that, in open-ended QA, negative samples primarily enhance the diversity and semantic relevance of generated responses, whereas positive samples contribute mainly to improving response quality. Building on this observation, we further conduct a systematic search over fixed asymmetric weightings of positive and negative samples, and find that the weight on negative samples governs the performance upper bound and the degree of exploration, while the weight on positive samples determines the convergence behavior and overall stability. These findings reveal that any fixed weighting scheme over positive and negative samples is inherently incapable of accommodating the reward distribution characteristic of open-ended QA.

Building upon the above observations, we further propose \textbf{EAPO}, an \textbf{E}ntropy-driven \textbf{A}daptive positive-negative sample weighting method for \textbf{P}olicy \textbf{O}ptimization tailored to open-ended QA scenarios. Specifically, EAPO adaptively modulates the weighting coefficient of positive samples during training according to the ratio between the current policy entropy and the initial entropy. During the entropy-decreasing phase, in which the policy progressively becomes more deterministic, EAPO automatically reduces the weight of positive samples so as to preserve the model's exploratory capacity. Conversely, during the entropy-increasing phase, where the policy turns overly stochastic, EAPO raises the weight of positive samples to reinforce response stability. Through such a dynamic balance, EAPO maintains the reliability of generated responses while effectively preventing entropy collapse. Extensive experiments on two publicly available open-ended medical QA datasets demonstrate that EAPO consistently outperforms baseline models employing fixed positive-negative sample weights in terms of both response diversity and stability, thereby offering a new perspective on the application of reinforcement learning to open-ended QA tasks.

Our contributions are summarized as follows:
\begin{itemize}
    \item We propose a reward‑mean‑based strategy for distinguishing positive and negative samples and systematically investigate the roles of these samples in reinforcement learning for open‑ended QA.

    \item We conduct a systematic search over fixed asymmetric weightings of positive and negative samples, and reveal that fixed weighting schemes are inherently unable to accommodate the reward distribution characteristic of open-ended QA.

    \item We propose EAPO, adaptively computes the weighting coefficients of positive samples based on the ratio of the current policy entropy to the initial entropy, thereby ensuring both response credibility and diversity in open-ended QA. We offers a new perspective on the application of reinforcement learning to open‑ended QA.

\end{itemize}


\section{Effects of Positive and Negative Samples in Open-Ended QA}

In this section, we first present the task formulation. We then introduce the proposed reward-mean-based strategy for distinguishing positive and negative samples in the open-ended QA reinforcement learning setting, together with the design of the reward function. Finally, we analyze the respective effects of positive and negative samples on reinforcement learning for open-ended QA.

\subsection{Task Formulation}

Given a prompt $x \in \mathcal{D}$, a language model policy $\pi_\theta$ generates a
response $y$. 
For each prompt $x$, we sample $N$ rollouts $\{y_i\}_{i=1}^N \sim \pi_\theta(\cdot \mid x)$
with corresponding rewards $\{r_i\}_{i=1}^N$. 
Unlike mathematical reasoning tasks where rewards are
binary, open-ended QA employs a continuous reward function $r(x, y)$ that
evaluates answer quality along multiple dimensions.
The training objective is to update
$\theta$ via policy gradient to increase the probability of higher-quality responses.

We follow previous work\citep{tang2025rethinking,zhu2026surprising} and decompose the open-ended QA reinforcement learning objective into two distinct learning paradigms: learning from correct rollouts and learning from incorrect rollouts. 
The overall objective can thereby be expressed as the sum of two sub-objectives:
\begin{equation}
\mathcal{L}(\theta) = w^{+}\mathcal{L}_{\mathrm{PSR}}(\theta) + w^{-}\mathcal{L}_{\mathrm{NSR}}(\theta),
\end{equation}
where $w^+$ and $w^-$ denote the weights of positive and negative samples, respectively. The two sub-objectives correspond to the two learning paradigms described above:
\begin{equation}
\mathcal{L}_{\mathrm{PSR}}(\theta) = -\mathbb{E}_{x \sim \mathcal{D}} \left[ \sum_{y:\, r(x,y) \geq \bar{r}_x} \pi_{\theta}(y \mid x) \right]
\end{equation}
\begin{equation}
\mathcal{L}_{\mathrm{NSR}}(\theta) = -\mathbb{E}_{x \sim \mathcal{D}} \left[ \sum_{y:\, r(x,y)< \bar{r}_x} -\pi_{\theta}(y \mid x) \right]
\end{equation}
where $\mathcal{L}_{\mathrm{PSR}}(\theta)$ denotes \textit{positive sample reinforcement} (PSR), and $\mathcal{L}_{\mathrm{NSR}}(\theta)$ denotes \textit{negative sample reinforcement} (NSR). $\bar{r}_x$ denotes the mean reward over the rollouts within a group, whose formal definition is provided in Subsection \ref{sec:partition}.



\subsection{Reward-Mean-Based Positive and Negative Sample Identification}
\label{sec:partition}

Unlike RLVR, open‑ended QA does not presuppose fixed ground‑truth correctness criteria. However, 
in GRPO \citep{guo2025deepseek}, the group-normalized advantage for rollout $i$ under prompt $x$ is:
\begin{equation}
  A_i = \frac{r_i - \bar{r}_x}{\sigma_x + \epsilon}, \quad
  \bar{r}_x = \frac{1}{N}\sum_{j=1}^N r_j
\end{equation}
$\sigma_{x}$ denotes the intra-group reward variance. The group mean $\bar{r}_x$ precisely serves as the critical boundary at which the direction of the reinforcement learning update reverses. Rollouts satisfying $r_i \geq \bar{r}_x$ are assigned a non-negative advantage $A_i \geq 0$ and are regarded as \textit{positive samples}, indicating that the model is encouraged to generate such trajectories with higher probability. Conversely, rollouts with $r_i < \bar{r}_x$ are assigned a negative advantage $A_i < 0$ and are treated as \textit{negative samples}, in which case the model is driven to reduce the likelihood of producing them.
Formally, we define:
\begin{equation}
  \mathcal{D}^+_x = \{i \mid r_i \geq \bar{r}_x\}, \quad
  \mathcal{D}^-_x = \{i \mid r_i < \bar{r}_x\}
\end{equation}
where $\mathcal{D}^+_x$ is the positive samples set and $\mathcal{D}^-_x$ is the negative samples set. The mean-based boundary is query-adaptive: it automatically calibrates to the current policy's output distribution for each prompt without any pre-defined correctness labels, making it particularly well-suited to open-ended QA, where no strict binary ground truth exists.

\subsection{Reward Design}

For the open-ended QA task, we design a composite reward combining four
complementary signals:
\begin{equation}
  \begin{split}
    r(x, y) = {}&\alpha_1 r_{\mathrm{fmt}}(y) + \alpha_2 r_{\mathrm{rouge}}(y, \hat{d}) \\
                &+ \alpha_3 r_{\mathrm{rerank}}(y, \hat{d}) + \alpha_4 r_{\mathrm{llm}}(x, y, \hat{d})
  \end{split}
\end{equation}
where $\hat{d}$ is the ground-truth response with weights $\alpha_1, \alpha_2, \alpha_3$ and $\alpha_4$.

\noindent\textbf{Format Reward} $r_{\mathrm{fmt}}(y) \in \{0,1\}$ checks whether the
response contains all required structural tags.

\noindent\textbf{Rouge-L Reward} $r_{\mathrm{rouge}}(y, \hat{d}) \in [0,1]$ computes
the character-level Rouge-L F1 score between the response and the
ground-truth, measuring lexical overlap.

\noindent\textbf{Reranker Reward} $r_{\mathrm{rerank}}(y, \hat{d}) \in [0,1]$ uses a
neural reranker to score the semantic relevance between the ground-truth $\hat{d}$ as
query and the generated response $y$ as document, capturing semantic similarity
beyond lexical overlap.

\noindent\textbf{LLM-as-a-Judge Reward} $r_{\mathrm{llm}}(x, y, \hat{d}) \in [0,1]$
evaluates the quality of the model's reasoning process within the
\texttt{<think>} tag via a large language model judge. The detailed prompt and evaluation dimensions are provided in Appendix \ref{appendix_prompt}.


\subsection{Experimental Setup}

\begin{figure*}[t]
    \centering
    \includegraphics[scale=0.3]{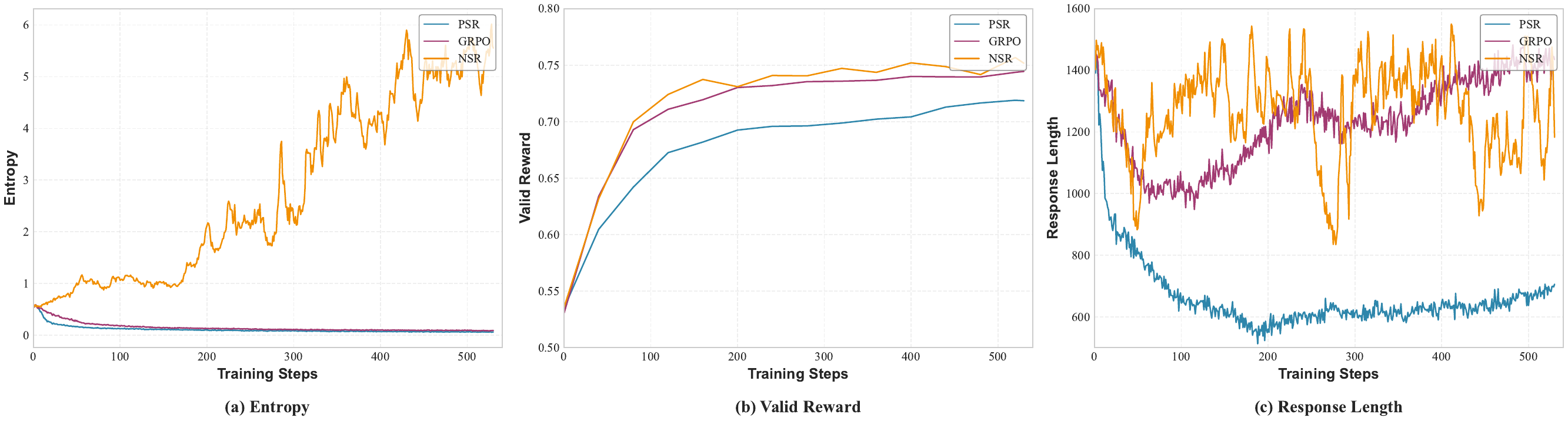}
    \caption{Training Dynamics on the RJUA dataset.
    }
    \label{fig:analysis}
\end{figure*}

We conduct experiments on Qwen3‑8B \citep{qwen3technicalreport}. Following the approach of W-REINFORCE \citep{zhu2026surprising}, we perform reinforcement learning separately using only positive samples and only negative samples for LLM, and include GRPO \citep{guo2025deepseek}, which utilizes both types of samples. 
The experimental setups are given in Subsection \ref{hyperaramter}. Details of the training and evaluation datasets are provided in Subsection \ref{datas}.
Evaluation methods are shown in Subsection \ref{evaluation}.

\subsection{Different Training Dynamics of Positive
and Negative Sample Reinforcement}

\textbf{Positive samples reduce entropy, while negative samples increase entropy.} 
As shown in Figure~\ref{fig:analysis}(a), PSR causes the model entropy to drop rapidly, whereas NSR maintains a higher entropy. PSR amplifies the logits of tokens in correct solutions, quickly converging to stable output patterns and yielding higher confidence on high-probability predictions. NSR, in contrast, suppresses the logits of tokens in incorrect solutions while indirectly lifting alternative tokens, thereby dispersing the probability distribution and preserving exploration diversity. This observation is consistent with the findings of W-REINFORCE~\citep{zhu2026surprising} and A3PO~\citep{tang2025rethinking} under the RLVR setting.

\textbf{Positive sample rewards improve slowly, while negative sample rewards improve quickly.}
As shown in Figure~\ref{fig:analysis}(b), PSR yields limited reward improvement. In contrast, NSR leads to faster reward gains, as NSR provide denser prohibitive signals that drive the policy away from high-error regions more rapidly. 

\textbf{Positive samples produce shorter responses, while negative samples yield longer ones.} 
As shown in Figure~\ref{fig:analysis}(c), PSR generate increasingly shorter responses, whereas NSR produce longer ones. This is because PSR rewards the shortest path to correct answers, leading to a substantial reduction in response length and potentially compromising reasoning depth in open-ended QA. NSR, by contrast, merely suppresses incorrect tokens without encouraging brevity, thereby allowing the model to explore longer reasoning chains.


\subsection{Multi-View Trade-offs of Positive and Negative Sample Reinforcement}

We evaluate the performance of PSR, NSR, and GRPO on Qwen3-8B across three evaluation dimensions: Rouge-L, Reranker, and LLM-as-a-Judge(LAAJ). The experimental results reveal a trade-off specific to open‑ended QA, as shown in Table \ref{tab:grpo-psr-nsr} and Table \ref{tab:rl_paradigm_metrics}. We take the CMD test set as a representative example for analysis and the conclusions drawn are equally applicable to the RJUA test set.

\begin{table}[t]
\centering
\small
\begin{tabular}{lcccc}
\toprule
Methods & Rouge-L & Reranker & LAAJ & Avg \\
\midrule
GRPO & 0.312 & 0.976 & \textbf{0.744} & \textbf{0.677} \\
PSR  & 0.293 & 0.940 & 0.738 & 0.657 \\
NSR  & \textbf{0.327} & \textbf{0.971} & 0.702 & 0.667 \\
\bottomrule
\end{tabular}
\caption{Performance comparison of different RL paradigms on RJUA test set. Avg: Average.}
\label{tab:grpo-psr-nsr}
\end{table}

\begin{table}[t]
\centering
\small
\begin{tabular}{lcccc}
\toprule
Methods & Rouge-L & Reranker & LAAJ & Avg \\
\midrule
GRPO & 0.186 & 0.783 & 0.649 & \textbf{0.539} \\
PSR & 0.147 & 0.688 & \textbf{0.783} & \textbf{0.539} \\
NSR & \textbf{0.194} & \textbf{0.792} & 0.544 & 0.510 \\
\bottomrule
\end{tabular}%
\caption{Performance comparison of different RL paradigms on CMD test set. Avg: Average.}
\label{tab:rl_paradigm_metrics}
\end{table}

\textbf{Positive samples improves answer quality but narrows the distribution.}
Compared with NSR, PSR yields a 0.239 (0.783 v.s. 0.544) improvement on the LAAJ metric, indicating that the responses generated by PSR are closer to high-quality answers. However, both the Rouge-L and Reranker scores decline, reflecting a narrowing of the output distribution under PSR and suggesting a risk of diversity collapse, whereby the model tends to replicate the stylistic patterns of positive samples.

\textbf{Negative samples enhances diversity and semantic relevance.} 
NSR achieves the best performance on both the ROUGE-L and Reranker metrics. This is because NSR steers the model away from low-quality answers, thereby expanding the generation space and encouraging the model to explore a broader answer distribution, which in turn helps prevent mode collapse.

PSR sacrifices exploration capability, while NSR compromises answer precision. By jointly incorporating positive and negative samples, GRPO achieves a more balanced performance across all three metrics, demonstrating the importance of the synergy between positive and negative samples in reinforcement training.

\section{The Impact of Asymmetric Weights in Open-ended QA Reinforcement Training}

\begin{figure*}[t]
    \centering
    \includegraphics[scale=0.3]{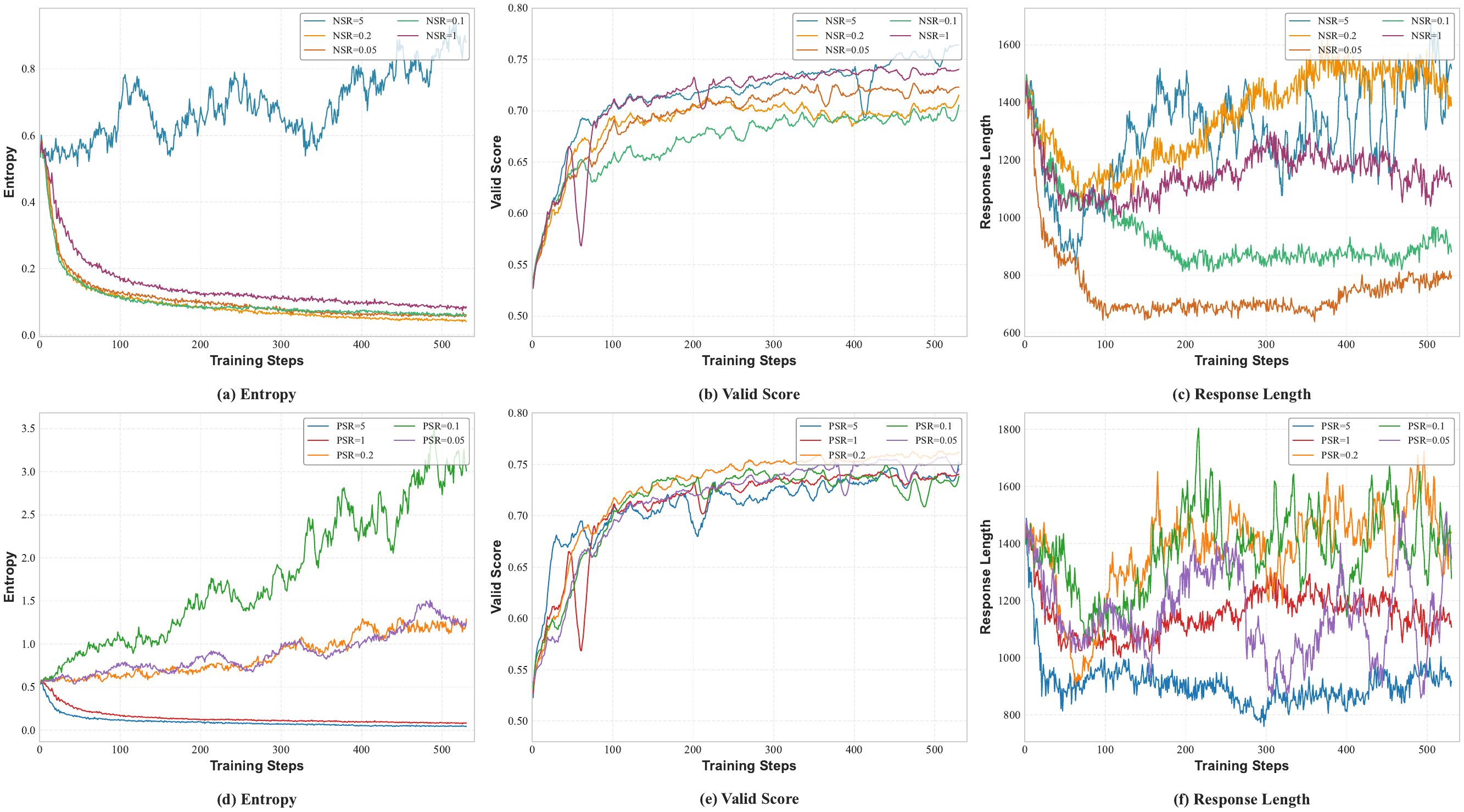}
    \caption{Training dynamics under asymmetric weighting on the RJUA dataset. The corresponding results on the CMD dataset are provided in Appendix \ref{appendix_asymmetric}.
    }
    \label{fig:analysis_v2_1}
\end{figure*}

Our previous analyses demonstrate that training with only one sample polarity impairs model performance, confirming the importance of both positive and negative samples in reinforcement learning for open-ended QA. In this section, we further investigate how adjusting the influence of each polarity affects the training process of reinforcement learning for open-ended QA.

\subsection{Training Dynamics of Asymmetric Weights for Positive and Negative Samples}

We systematically investigate the impact of weight imbalance between PSR and NSR on the training process. The experimental design follows a controlled ablation: with the positive sample weight $w^+ = 1.0$, the negative sample weight $w^- \in \{0.05,0.1,0.2,1,5\} $, as shown in the first row of Figure \ref{fig:analysis_v2_1}; with the negative sample weight $w^- = 1.0$, the positive sample weight $w^+ \in \{0.05,0.1,0.2,1,5\} $, as shown in the second row of Figure \ref{fig:analysis_v2_1}.

\textbf{A larger negative sample weight leads to higher entropy and longer response lengths.} 
When the negative sample weight is too small ($\le 0.2$), the training procedure degenerates toward the PSR, leading to a rapid decline in entropy and a shortening of response length. A larger negative sample weight more effectively prevents policy collapse. Notably, with the negative sample weight set to 5, the model attains the highest reward, yet exhibits substantial fluctuations in the later stages of training. Hence, simply enlarging the negative sample weight to improve performance suffers from diminishing marginal returns and degraded training stability.

\textbf{The larger the positive sample weight, the lower the entropy and the shorter the response length.} 
A larger positive sample weight drives the entropy closer to collapse and yields shorter responses. When the positive sample weight is set to 5, the model entropy collapses almost completely and the response length reaches its minimum, indicating that the output distribution becomes increasingly concentrated and the generation increasingly conservative.

\textbf{Negative sample weight determines the performance upper bound and exploration, while positive sample weight determines the convergence behavior and stability.} 
Notably, over the two-orders-of-magnitude range from 0.05 to 5, the reward variation (the gap between the maximum and minimum) induced by the negative sample weight is substantially larger than that induced by the positive sample weight. This indicates that the negative sample weight dominates the performance upper bound and exploration capability, whereas the positive sample weight governs convergence behavior and training stability. Consequently, the key to reinforcement learning for open-ended QA lies in the effective exploitation of negative samples. The underlying reason is that the generation space of open-ended QA is exceedingly vast, and positive samples can only cover a limited set of patterns. The model primarily acquires generalization capability by learning "what should not be generated".

\subsection{Relationship Between the P/N Sample Weight Ratio and Performance}
\label{sec_relation}


\begin{table}[h]
\centering
\small
\resizebox{0.5\textwidth}{!}{%
\begin{tabular}{lccccc}
\toprule
Methods & $w^+$ & $w^-$ & $w^+/w^-$ & Rouge-L & Reranker \\
\midrule
PSR & 1 & 0 & $\infty$ & 0.293 & 0.940 \\
NSR & 0 & 1 & 0 & 0.327 & 0.971 \\
GRPO & 1 & 1 & 1 & 0.301 & 0.970 \\
\midrule
\multirow{4}{*}{P/N $\downarrow$ }
& 1 & 0.05 & 20 & 0.299 & 0.933 \\
& 1 & 0.1 & 10 & 0.232 & 0.922 \\
& 1 & 0.2 & 5 & 0.274 & 0.931 \\
& 1 & 5 & 0.2 & \textbf{0.335} & \textbf{0.976} \\
\cdashline{1-6}
\multirow{4}{*}{P/N $\uparrow$ }
& 0.05 & 1 & 0.05 & 0.263 & 0.903 \\
& 0.1 & 1 & 0.1 & 0.322 & 0.958 \\
& 0.2 & 1 & 0.2 & \textbf{0.337} & \textbf{0.983} \\
& 5 & 1 & 5 & 0.302 & 0.970 \\
\bottomrule
\end{tabular}
}
\caption{Performance under different positive/negative reward weight configurations.}
\label{tab:weight_ablation_v2}
\end{table}

We evaluate the performance under different positive-to-negative sample weight ratios based on Qwen3-8B on the RJUA test set, and the evaluation results are shown in Table \ref{tab:weight_ablation_v2}. The experimental results on CMD are provided in Appendix \ref{relation}.


\textbf{Dominance of positive sample weights leads to mode collapse, whereas dominance of negative samples enhances diversity.} 
When the positive sample weight is fixed and the negative sample weight is progressively increased, the exploratory capability of the model gradually takes precedence, with the Rouge-L and Reranker scores rising to 0.335 and 0.976, respectively, markedly surpassing NSR. This indicates that retaining a small portion of positive samples, combined with the exploration driven by negative samples, yields superior performance compared with NSR. Conversely, when the negative sample weight is fixed and the positive sample weight is progressively increased, the model performance exhibits a rise-then-fall pattern: when the positive-sample weight reaches 5, the performance drops, suggesting that an excessively strong positive sample influence compresses the output distribution and drives the training back into a collapse regime.

\textbf{The optimal performance is achieved when weight ratio of positive and negative sample is 1:5.} 
An interesting observation is that the optimal weight ratio between positive and negative samples is not 1:1, but rather approximately 1:5, with negative samples being dominant. Two independent weight-sweep experiments both converge to this same ratio. This suggests that, in the open-domain QA setting, the effective signal density of the negative sample gradient is lower than that of the positive sample gradient, and consequently a roughly fivefold weight is required to balance against the positive samples.

\section{Entropy-Driven Adaptive Positive-Negative Sample Weighting }

Based on the preceding analyses, in reinforcement learning for open-ended QA, any fixed ratio between the positive- and negative-sample weights corresponds to an entropy trajectory of a predetermined direction, which is incapable of perceiving the actual demands of the current training stage. To address this limitation, we propose EAPO, an entropy-driven adaptive positive-negative sample weighting strategy for policy optimization. 

\subsection{Method}

EAPO dynamically adjusts the weight of positive samples by leveraging the ratio between the current policy entropy $H_t$ and the initial entropy $H_0$, thereby adaptively balancing exploration and exploitation throughout the entire training process. We provide a detailed derivation of the formula in Appendix \ref{appendix_formula}.
The dynamic positive sample
weight is defined as:
\begin{equation}
  w_t^+ = \mathrm{clip}\!\left(w_0 \cdot \frac{H_t}{H_0},\ w_{\min},\ w_{\max}\right)
  \label{eq:dynamic_weight}
\end{equation}
where $w_0$ is the base positive weight and
$[w_{\min}, w_{\max}]$ bounds the weight within a safe range. The negative sample weight $w^-$ is fixed at as the reference
baseline. We use the ratio $H_t / H_0$ rather than the absolute difference to ensure
scale-invariance across base models with different initial entropy levels.

Given the dynamic weight, the weighted advantage for rollout $y_i$ is:
\begin{equation}
  \hat{A}_i =
  \begin{cases}
    w_t^+ \cdot A_i & i \in \mathcal{D}^+_x \\[2pt]
    w^-  \cdot A_i  & i \in \mathcal{D}^-_x
  \end{cases}
\end{equation}


The dynamic behavior of EAPO can be interpreted as follows. During the entropy-decreasing phase ($H_t < H_0$), positive sample reinforcement becomes dominant and the policy over-concentrates; in this case, setting $w_t^+ < w_0$ attenuates the strength of positive reinforcement, allowing negative samples to counteract the tendency toward over-convergence. Conversely, during the entropy-increasing phase ($H_t > H_0$), the exploratory force prevails and the policy becomes overly diffuse; here, setting $w_t^+ > w_0$ amplifies positive reinforcement, thereby guiding the policy toward more confident outputs.


\begin{figure*}[t]
    \centering
    \includegraphics[scale=0.3]{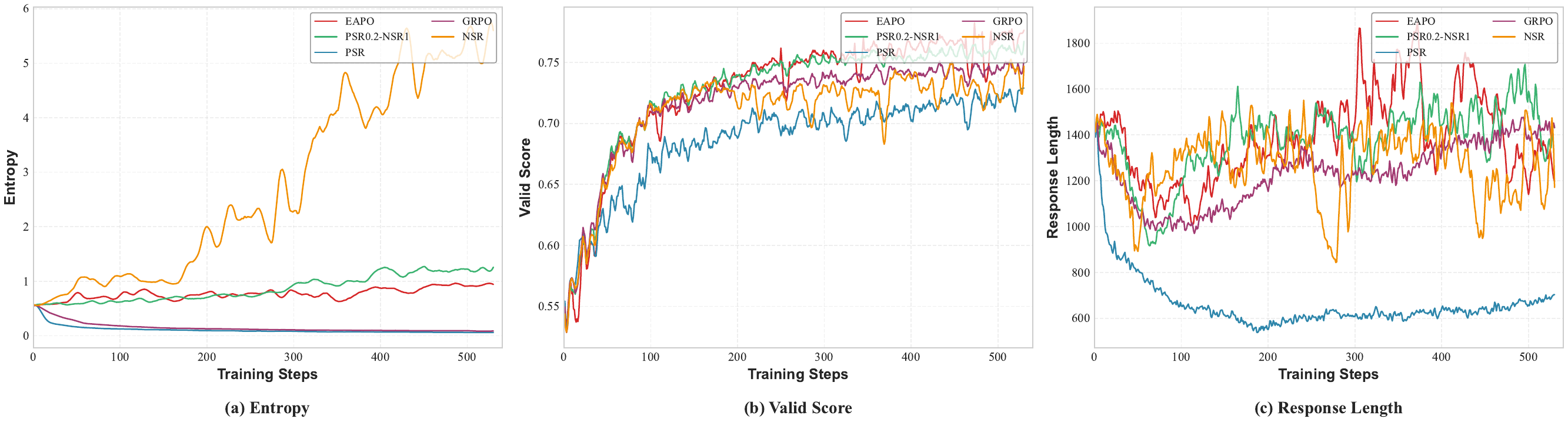}
    \caption{Training dynamics of EAPO against various baseline methods on the RJUA dataset.
    }
    \label{fig:analysis_v3}
\end{figure*}

The training objective of EAPO is as follows:

\begin{align}
\mathcal{J}(\theta) &= \mathbb{E}\bigg[ \frac{1}{N} \sum_{i=1}^{N} \frac{1}{|y_i|} \sum_{t=1}^{|y_i|} \min\!\big( r_t \hat{A}_{i,t},\; \notag \\
&\quad \text{clip}(r_t, 1{-}\epsilon, 1{+}\epsilon)\, \hat{A}_{i,t} \big) - \beta\, D_{\text{KL}}[\pi_\theta \| \pi_{\text{ref}}] \bigg]
\label{eq:grpo}
\end{align}

$r_t = \frac{\pi_\theta(y_{i,t} \mid x, y_{i,<t})}{\pi_{\theta_{\text{old}}}(y_{i,t} \mid x, y_{i,<t})}$ is the importance ratio. $\beta$ is the KL divergence coefficient, and the KL divergence penalty term is added to constrain policy shift.

  

\section{Experiments and Results}
\label{hyperaramter}
\subsection{Experimental Settings}
We select the following methods as baselines:
\textbf{GRPO}: both $w^+$ and $w^-$ are set to 1.0.
\textbf{W-REINFORCE}: fixed asymmetric weights, setting $w^+$ to 0.1.
\textbf{PSR}: training with positive samples only with $w^+=1.0, w^-=0.$
\textbf{NSR}: training with negative samples only with $w^+=0, w^-=1.0.$
All experiments are conducted using Qwen3-8B as the backbone.

We set the global training batch size to $32$. Both the maximum prompt length and the maximum response length are set to $2048$ tokens. The model is trained for a total of $10$ epochs with a learning rate of $1e^{-6}$ and a KL coefficient $\beta=0.001$. clipping range $\epsilon=0.2$.
Default hyperparameters for EAPO: $w_0=0.2, w_{min}=0, w_{max}=2.0$. During the rollout stage, $N = 4$ trajectories are sampled per prompt. All experiments are conducted on $4{\times}$NVIDIA H800 GPUs. We adopt DeepSeek-V3.2\citep{deepseekai2025deepseekv32} as the evaluator to score the model's reasoning process under the LLM-as-a-Judge protocol.

Details of the training data and evaluation protocol are provided in Appendix \ref{appendix_data_evaluation}.


\subsection{Main Results}

Figure \ref{fig:analysis_v3} compares the training dynamics of EAPO against various baseline methods. PSR and GRPO suffer from entropy collapse, whereas NSR and PSR0.2-NSR1 exhibit entropy explosion. In contrast, EAPO maintains a stable entropy level and confines the response length within a reasonable range. This demonstrates that EAPO is able to consistently strike a balance between exploration and exploitation, yielding a markedly more stable reinforcement learning training trajectory.

\begin{table}[t]
\centering
\small
\resizebox{0.5\textwidth}{!}{%
\begin{tabular}{lccccc}
\toprule
Methods & Rouge-L & Reranker & RL@8 & RR@8 & Avg \\
\midrule
GRPO & 0.312 & \textbf{0.976} & 0.344 & 0.987 & 0.655 \\
PSR & 0.293 & 0.940 & 0.327 & 0.979 & 0.635 \\
NSR & 0.327 & 0.971 & 0.361 & 0.990 & 0.662 \\
W-REINFORCE & 0.322 & 0.958 & \textbf{0.366} & \textbf{0.998} & 0.661 \\
\hline
EAPO & \textbf{0.330} & 0.973 & 0.360 & 0.995 & \textbf{0.664} \\
\bottomrule
\end{tabular}%
}
\caption{Performance comparison of different reinforcement learning methods. RL@8 denotes the best Rouge-L score among $8$ inference samples, while RR@8 denotes the best Reranker score among $8$ inference samples.}
\label{tab:rl_method_comparison}
\end{table}

The main results on RJUA test set are shown in Table \ref{tab:rl_method_comparison}, where @k denotes the best result after k inference runs of the model. Appendix \ref{appendix_generalization} provides a performance comparison of different methods built upon Qwen2.5-7B-Instruct\citep{qwen2025qwen25technicalreport}. We draw the following conclusions:

\textbf{In open‑domain question answering, positive samples lack diversity, while negative samples suffer from insufficient stability.} PSR achieves the lowest performance across all metrics, indicating that simply reinforcing correct answers in open‑domain QA traps the model into repeating already mastered patterns, thereby losing diversity. NSR expands output diversity and is very friendly to single‑generation quality (Rouge‑L, Reranker). However, its RR@8 is lower than those of W‑REINFORCE and EAPO, suggesting that with only punishment and no reward, the model struggles to stably converge to high‑quality answers across multiple sampling runs, resulting in excessive exploration but insufficient exploitation.

\textbf{Fixed weights cannot balance single‑generation quality and sampling upper bound simultaneously.} W‑REINFORCE achieves the best RL@8 and RR@8, but its Rouge‑L and Reranker scores are significantly lower than those of NSR and EAPO. This indicates that a fixed ratio of low‑weight positive samples to high‑weight negative samples can yield high‑scoring answers under multiple sampling runs, but the average quality of single generations declines. Fixed weights cannot balance single‑generation quality and sampling upper bound, and the weights require manual tuning, incurring high transfer costs.
Appendix \ref{appendix_weight} presents the effects of different fixed weighting configurations, together with a stability analysis of EAPO's performance.

\textbf{Entropy‑aware adaptive weights are robust.} EAPO approaches the optimum on every metric and achieves the best overall performance. 
This also suggests that the key to RL for open‑domain QA is not merely rewarding correct answers, but rewarding them without collapsing the distribution. EAPO performs online interpolation between the extreme strategies of PSR and NSR throughout training, thus achieving the most robust comprehensive performance.


\section{Related Work}

\textbf{Reinforcement Learning for Large Reasoning Models.} Reinforcement learning has become a central paradigm for improving the reasoning ability of large language models. Early alignment methods commonly optimize language model policies with PPO-style objectives and KL constraints~\citep{schulman2017proximal}. More recent reasoning-oriented methods adopt reinforcement learning with verifiable rewards (RLVR)\citep{chen2025exploration,xu2026you,xie2025unlocking,zheng2025group,wang2025aspo}, where models generate multiple candidate reasoning trajectories and receive outcome-based rewards. DeepSeekMath~\citep{shao2024deepseekmath} introduces GRPO, which estimates advantages from groups of sampled rollouts and removes the need for a separate value model. DeepSeek-R1~\citep{guo2025deepseek} further demonstrates that large-scale RL can elicit complex reasoning behaviors from base models. Follow-up systems such as DAPO~\citep{yu2026dapo} improve the stability and scalability of long-CoT RL through algorithmic and system-level refinements. 

\textbf{Positive and Negative Sample Signals in RLVR.} 
Recent studies\citep{zhu2026surprising,tang2025rethinking,arnal2026asymmetric} have begun to revisit the role of sample polarity in RLVR. Rather than treating incorrect or low-reward rollouts as uninformative failures, negative samples can provide useful signals for exploration. \citet{zhu2026surprising} show that negative reinforcement can be surprisingly effective for LLM reasoning, preserving diversity and improving high-$k$ performance. \citet{tang2025rethinking} further analyze positive and negative samples in RLVR and show that positive samples tend to sharpen existing correct reasoning patterns, whereas negative samples encourage exploration of alternative reasoning paths.
However, these methods have been evaluated only in RLVR settings, and their conclusions have yet to be validated in open-ended question answering scenarios.

\textbf{Entropy-guided Adaptive Policy Optimization.}
Entropy has long been used as a proxy for policy exploration in reinforcement learning. Maximum-entropy RL encourages policies to maintain stochasticity and improves exploration robustness~\citep{ziebart2008maximum,haarnoja2018soft,tian2026learning}, while PPO-style objectives often include entropy bonuses or KL penalties to prevent unstable policy updates~\citep{schulman2017proximal}. In LLM reinforcement learning, 
entropy collapse often corresponds to shortened outputs, reduced diversity, and premature convergence, whereas excessive entropy may lead to unstable or low-confidence generations. 
We propose EAPO, which adaptively determines the weighting of positive samples based on the ratio between the current policy entropy and the initial reference entropy, dynamically balancing exploration and exploitation in open-ened QA.

\section{Conclusion}
We propose a reward-mean-based criterion to distinguish positive and negative samples, and investigate their respective contributions to policy optimization for open-ended QA. We further propose EAPO, which adaptively determines the weighting coefficient of positive samples based on the ratio between the current policy entropy and the initial reference entropy. This entropy-driven mechanism dynamically balances exploration and exploitation in open-ended QA. 

\section*{Limitations}

This work is subject to several limitations. First, EAPO is validated only on open-ended medical question answering tasks with backbones of up to 8B parameters; its effectiveness on substantially larger models, non-medical domains, and multi-modal scenarios remains to be investigated. Second, the positive/negative sample partition relies on a reward-mean-based criterion, which is sensitive to outliers and to the absolute scale of the reward function; alternative strategies such as quantile-based or learnable thresholds have not yet been systematically explored. Third, we adopt LLM-as-a-Judge to evaluate reasoning quality, which may inherit biases from the underlying judge model. 
All public models and baselines used in this work are employed under their original licenses or service terms, while proprietary large language models are accessed through their official services. Future work will focus on scaling EAPO to larger and multi-modal foundation models, exploring more robust anchor and sample-partitioning mechanisms, and further validating its effectiveness on a broader range of open-ended question answering benchmarks.




\bibliography{custom}
\clearpage

\appendix

\section{Prompt and Evaluation Dimensions}
\label{appendix_prompt}

\subsection{Prompt}
We conduct experiments on two open-ended medical question-answering datasets, RJUA and CMD, with the detailed description on Appendix \ref{datas}. The corresponding prompts presented as follows.

\begin{tcolorbox}[promptbox, title=Prompt on RJUA]



You are an expert physician with comprehensive clinical expertise in Chinese urology and andrology.
Please read the patient's consultation carefully and respond according to the following steps:
1. Perform clinical diagnostic reasoning within the \textbf{<think>} and \textbf{</think>} tags. The reasoning should encompass information acquisition and integration, disease hypothesis generation, hypothesis verification and exclusion, and clinical reasoning strategy.
2. Provide the diagnostic conclusion together with recommendations for further diagnostic workup or treatment within the \textbf{<advice>} and \textbf{</advice>} tags.\\

\medskip
Patient consultation: \{question\}
\end{tcolorbox}

\begin{tcolorbox}[promptbox, title=Prompt on CMD]



You are an expert physician with comprehensive clinical expertise across all medical specialties in China.
Please read the patient's consultation carefully and respond according to the following steps:
1. Perform clinical diagnostic reasoning within the \textbf{<think>} and \textbf{</think>} tags. The reasoning should encompass information acquisition and integration, disease hypothesis generation, hypothesis verification and exclusion, and clinical reasoning strategy.
2. Provide the diagnostic conclusion together with recommendations for further diagnostic workup or treatment within the \textbf{<advice>} and \textbf{</advice>} tags.\\

\medskip
Patient consultation: \{question\}
\end{tcolorbox}

\subsection{Evaluation Dimensions on LLM-as-a-Judge}

We adopt an LLM-as-a-Judge paradigm to score the model's reasoning content in \textit{<think>} tag during reinforcement training and the corresponding judge prompt is presented as follows:

\begin{tcolorbox}[promptbox, title=Prompt on LLM-as-a-Judge for <think>]



You are a senior clinical medicine expert and a specialist in medical-education assessment. Your task is to evaluate the quality of the **internal reasoning process** of a medical consultation system.\\

\medskip
Note: You are required to assess only the model's reasoning process (the content within the <think> tags); the final response delivered to the patient is not subject to evaluation.\\

\medskip
\textbf{[Patient Consultation]}\\
\{query\}\\

\medskip
\textbf{[Reference Answer (Ground Truth)]}\\
\{gold\_answer\}\\

\medskip
\textbf{[Model's Reasoning Process (to be evaluated)]}
\{think\_content\}\\

\medskip
Please rate the model's reasoning process along the following four dimensions, each on a 100-point scale.\\

\medskip
\textbf{[Dimension 1: Logical Coherence of Clinical Reasoning — Weight 35\%] }\\ \medskip
\textbf{[[Dimension 2: Accuracy of Medical Knowledge — Weight 30\%]} \\
\medskip
\textbf{[[Dimension 3: Adequacy of Differential Diagnosis — Weight 20\%]} \\
\medskip
\textbf{[[Dimension 4: Depth and Organization of Reasoning — Weight 15\%]} \\

\medskip
Overall Score = Logical Coherence × 0.35 + Medical Knowledge Accuracy × 0.30 + Differential Diagnosis Adequacy × 0.20 + Depth and Organization × 0.15
\end{tcolorbox}

The prompt used to evaluate model-generated answers under the LLM-as-a-Judge paradigm is presented as follows:

\begin{tcolorbox}[promptbox, title=Prompt on LLM-as-a-Judge for Response]



You are a senior clinical medicine expert and a specialist in medical-education assessment. Your task is to perform a multi-dimensional evaluation of the response produced by a medical consultation system.\\

\medskip
\textbf{[Patient Consultation Query]}\\
\{query\}\\

\medskip
\textbf{[Reference Answer (Ground Truth)]}\\
\{gold\_diagnosis\}\\

\medskip
\textbf{[Model Response (To Be Evaluated)]}
\{response\_content\}\\

\medskip
Please rate the model response along the following five dimensions, with each dimension scored on a 100 scale.\\

\medskip
\textbf{[Dimension 1: Medical Factual Accuracy — Weight 30\%] }\\ \medskip
\textbf{[[Dimension 2: Diagnostic Completeness — Weight 25\%]} \\
\medskip
\textbf{[[Dimension 3: Clinical Safety — Weight 25\%]} \\
\medskip
\textbf{[[Dimension 4: Reasoning Logic and Interpretability — Weight 10\%]} \\
\medskip
\textbf{[[Dimension 5: Expression Clarity and Empathy — Weight 10\%]} \\

\medskip
Overall Score = Medical Factual Accuracy $\times$ 0.30 + Diagnostic Completeness $\times$ 0.25 + Clinical Safety $\times$ 0.25 + Reasoning Logic $\times$ 0.10 + Expression Clarity $\times$ 0.10
\end{tcolorbox}

\section{The Impact of Asymmetric Weights in Open-ended QA Reinforcement Training}
\label{appendix_asymmetric}

\begin{figure*}[t]
    \centering
    \includegraphics[scale=0.3]{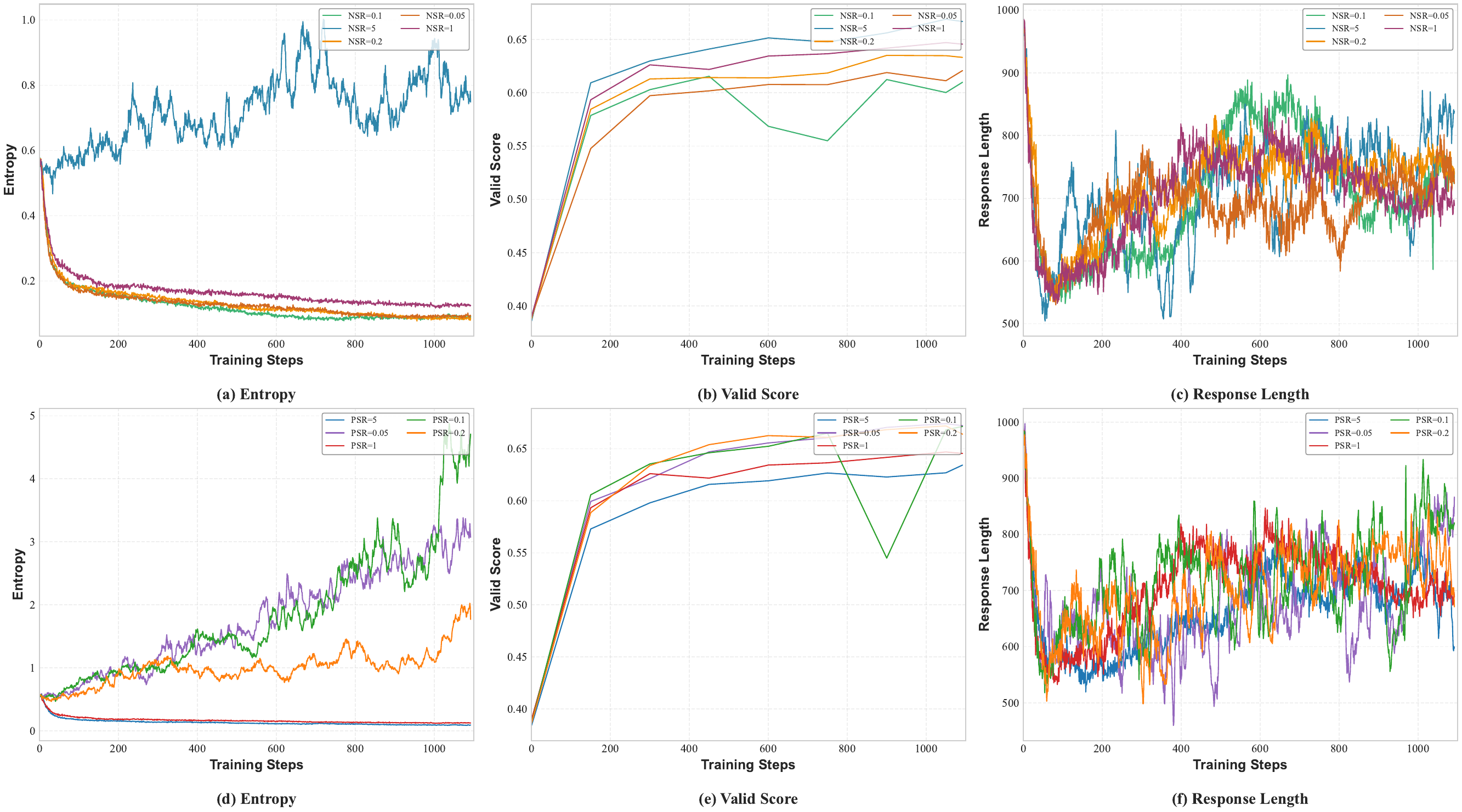}
    \caption{Training dynamics under asymmetric weighting on the CMD dataset.
    }
    \label{fig:analysis_v2}
\end{figure*}

Figure \ref{fig:analysis_v2} reports the training dynamics on the CMD dataset under different asymmetric weighting configurations, where the upper row varies $w^{-}$ with $w^{+}=1$ fixed (NSR series), and the lower row varies $w^{+}$ with $w^{-}=1$ fixed (PSR series).

Across all configurations, no fixed $(w^{+}, w^{-})$ setting jointly stabilizes entropy variance, response length, and validation score, consistent with the failure modes observed on RJUA. These results further confirm that static asymmetric weighting is fundamentally incapable of accommodating the reward distribution characteristic of open-ended QA.

\section{Relation Between the P/N Sample Weight Ratio and Performance}
\label{relation}

Table \ref{tab:weight_ablation} reports the performance of various $(w^{+}, w^{-})$ configurations on the CMD dataset.

It can be observed that NSR consistently outperforms PSR in overall performance, and that the model's performance progressively degrades as the positive-sample weight increases. These observations are consistent with the experimental results in Section \ref{sec_relation}, further corroborating that the dominance of positive-sample weights tends to induce mode collapse, whereas the dominance of negative-sample weights promotes generation diversity.

\begin{table}[t]
\centering
\small
\resizebox{0.5\textwidth}{!}{%
\begin{tabular}{lccccc}
\toprule
Methods & $w^{+}$ & $w^{-}$ & $w^{+}/w^{-}$ & Rouge-L & Reranker \\
\midrule
PSR  & 1 & 0 & $\infty$ & 0.147 & 0.688 \\
NSR  & 0 & 1 & 0        & 0.194 & 0.792 \\
GRPO & 1 & 1 & 1        & 0.184 & 0.787 \\
\midrule
\multirow{4}{*}{P/N $\downarrow$}
     & 1 & 0.05 & 20  & 0.160 & 0.749 \\
     & 1 & 0.1  & 10  & 0.158 & 0.724 \\
     & 1 & 0.2  & 5   & 0.170 & 0.763 \\
     & 1 & 5    & 0.2 & \textbf{0.187} & \textbf{0.793} \\
\midrule
\multirow{4}{*}{P/N $\uparrow$}
     & 0.05 & 1 & 0.05 & \textbf{0.194} & 0.778 \\
     & 0.1  & 1 & 0.1  & 0.185 & \textbf{0.808} \\
     & 0.2  & 1 & 0.2  & 0.186 & 0.754 \\
     & 5    & 1 & 5    & 0.173 & 0.752 \\
\bottomrule
\end{tabular}
}
\caption{Performance under different positive/negative reward weight configurations. Benchmark: CMD.}
\label{tab:weight_ablation}
\end{table}

\section{Formula Derivation}
\label{appendix_formula}

Following the intuition that the sample weight should be adjusted in proportion to the entropy variation between consecutive training steps, we define the recursive update rule as
\begin{equation}
w_t \;=\; w_{t-1} \cdot \frac{H_t}{H_{t-1}}
\label{eq:recursive}
\end{equation}
where $H_t$ denotes the rollout entropy at training step $t$.
Unrolling the recursion in Eq.~\eqref{eq:recursive} yields
\begin{equation}
\begin{aligned}
w_t
&= w_{t-1}\cdot \frac{H_t}{H_{t-1}}
   = w_{t-2}\cdot \frac{H_{t-1}}{H_{t-2}}\cdot \frac{H_t}{H_{t-1}} \\
&= \cdots
   = w_0 \cdot \frac{H_1}{H_0} \cdot \frac{H_2}{H_1} \cdots \frac{H_t}{H_{t-1}}
\end{aligned}
\end{equation}

Since the intermediate entropy terms cancel pairwise (telescoping product), the recursion collapses into the closed-form expression
\begin{equation}
\boxed{\,w_t \;=\; w_0 \cdot \dfrac{H_t}{H_0}\,}
\label{eq:closed-form}
\end{equation}

where the symbols are defined as follows:
\begin{itemize}
    \item $w_0$: the initial PSR weight, which serves as the only hyperparameter of the mechanism;
    \item $H_0$: the rollout entropy recorded at the first training step (fixed thereafter);
    \item $H_t$: the rollout entropy at the current step $t$;
    \item Weight clipping: $w_t \in [0, 2]$ to prevent unbounded amplification.
\end{itemize}

The closed form in Eq.~\eqref{eq:closed-form} reveals that, despite its recursive appearance, the entropy-driven weighting scheme is mathematically equivalent to a single anchor-based ratio between the current entropy $H_t$ and the initial reference entropy $H_0$. This formulation eliminates error accumulation across steps and provides a stable, hyperparameter-light reweighting mechanism.

\section{Datasets and Evaluation Methods}
\label{appendix_data_evaluation}
\subsection{Datasets}
\label{datas}

We evaluate our approach on two open‑domain medical question answering datasets:
CMD \citep{xu2023cmd}: A Chinese medical dialogue dataset. We take the subset of the top five departments, covering Internal Medicine, Surgery, Obstetrics and Gynecology, Pediatrics, and Dermatology. The training/test split is 9:1.
RJUA \citep{lyu2023rjuaqa}: A Chinese urological specialty diagnostic QA dataset, containing complete medical records and reference diagnoses.
Both datasets are open‑ended generation tasks with no single ground‑truth answer; answer quality must be evaluated from multiple dimensions.

\begin{table}[t]
\centering
\small
\setlength{\tabcolsep}{6pt}
\resizebox{0.5\textwidth}{!}{%
\begin{tabular}{lccccc}
\toprule
Methods & Rouge-L & Reranker & RL@8 & RR@8 & Avg \\
\midrule
GRPO & 0.312 & 0.977 & 0.344 & 0.987 & 0.655 \\
PSR  & 0.293 & 0.940 & 0.327 & 0.979 & 0.635 \\
NSR  & 0.327 & 0.971 & 0.361 & 0.990 & 0.662 \\
\midrule
\multicolumn{6}{l}{\textit{W-REINFORCE}} \\
\midrule
$w^{+}=0.05,\ w^{-}=1$ & 0.263 & 0.903 & 0.346 & 0.996 & 0.627 \\
$w^{+}=0.1,\ w^{-}=1$  & 0.322 & 0.958 & 0.366 & \textbf{0.998} & 0.661 \\
$w^{+}=0.2,\ w^{-}=1$  & \textbf{0.337} & \textbf{0.983} & \textbf{0.372} & 0.997 & \textbf{0.672} \\
$w^{+}=1,\ w^{-}=1$    & 0.301 & 0.970 & 0.326 & 0.994 & 0.648 \\
$w^{+}=5,\ w^{-}=1$    & 0.302 & 0.970 & 0.335 & 0.989 & 0.649 \\
$w^{+}=1,\ w^{-}=0.05$ & 0.299 & 0.933 & 0.324 & 0.983 & 0.635 \\
$w^{+}=1,\ w^{-}=0.1$  & 0.232 & 0.922 & 0.264 & 0.973 & 0.598 \\
$w^{+}=1,\ w^{-}=0.2$  & 0.274 & 0.931 & 0.288 & 0.978 & 0.618 \\
$w^{+}=1,\ w^{-}=5$    & 0.335 & 0.976 & 0.371 & 0.995 & 0.669 \\
\midrule
\textbf{EAPO} & 0.330 & 0.973 & 0.360 & 0.995 & 0.664 \\
\bottomrule
\end{tabular}
}
\caption{Effect of fixed asymmetric weights ($w^{+}, w^{-}$) within W-REINFORCE compared with GRPO, PSR, NSR, and our proposed EAPO across Rouge-L, Reranker, Best-RL@8, and Best-RR@8 metrics. Benchmark: RJUA.}
\label{tab:fixed-weight-ablation}
\end{table}

\subsection{Evaluation Methods}
\label{evaluation}

We report the following metrics:
\textbf{ROUGE‑L}: measures lexical coverage and diversity.
\textbf{Reranker}: measures semantic relevance.
\textbf{LLM‑as-a-Judge(LAAJ)}: measures reasoning chain quality and clinical credibility.

\begin{table}[t]
\centering
\small
\setlength{\tabcolsep}{6pt}
\resizebox{0.5\textwidth}{!}{%
\begin{tabular}{lccccc}
\toprule
Methods & Rouge-L & Reranker & RL@8 & RR@8 & Avg \\
\midrule
GRPO & 0.186 & 0.783 & 0.206 & 0.838 & 0.503 \\
PSR  & 0.147 & 0.688 & 0.173 & 0.821 & 0.457 \\
NSR  & \textbf{0.194} & 0.792 & 0.220 & \textbf{0.879} & \textbf{0.521} \\
\midrule
\multicolumn{6}{l}{\textit{W-REINFORCE}} \\
\midrule
$w^{+}=0.05,\ w^{-}=1$ & 0.194 & 0.778 & \textbf{0.220} & 0.874 & 0.517 \\
$w^{+}=0.1,\ w^{-}=1$  & 0.185 & \textbf{0.808} & 0.207 & 0.871 & 0.518 \\
$w^{+}=0.2,\ w^{-}=1$  & 0.186 & 0.754 & 0.216 & 0.876 & 0.508 \\
$w^{+}=1,\ w^{-}=1$    & 0.184 & 0.787 & 0.207 & 0.852 & 0.508 \\
$w^{+}=5,\ w^{-}=1$    & 0.173 & 0.752 & 0.199 & 0.838 & 0.490 \\
$w^{+}=1,\ w^{-}=0.05$ & 0.160 & 0.749 & 0.184 & 0.827 & 0.480 \\
$w^{+}=1,\ w^{-}=0.1$  & 0.158 & 0.724 & 0.183 & 0.806 & 0.467 \\
$w^{+}=1,\ w^{-}=0.2$  & 0.170 & 0.763 & 0.195 & 0.852 & 0.495 \\
$w^{+}=1,\ w^{-}=5$    & 0.187 & 0.793 & 0.210 & 0.862 & 0.513 \\
\midrule
\textbf{EAPO} & 0.187 & 0.798 & 0.213 & 0.864 & 0.516 \\
\bottomrule
\end{tabular}
}
\caption{Effect of fixed asymmetric weights ($w^{+}, w^{-}$) within W-REINFORCE compared with GRPO, PSR, NSR, and our proposed EAPO across Rouge-L, Reranker, Best-RL@8, and Best-RR@8 metrics. Benchmark: CMD.}
\label{tab:fixed-weight-ablation-cmd}
\end{table}



\begin{table}[ht]
\centering
\small
\renewcommand{\arraystretch}{1.15}
\setlength{\tabcolsep}{6pt}
\resizebox{0.5\textwidth}{!}{%
\begin{tabular}{lcccc}
\toprule
\multirow{2}{*}{Methods} & \multicolumn{2}{c}{RUJA} & \multicolumn{2}{c}{CMD} \\
\cmidrule(lr){2-3} \cmidrule(lr){4-5}
 & Rouge-L & Reranker & Rouge-L & Reranker \\
\midrule
\multicolumn{5}{l}{W-REINFORCE} \\
\quad $w^{+}=0.05$ & 0.263 & 0.903 & 0.194 & 0.778 \\
\quad $w^{+}=0.1$  & 0.322 & 0.958 & 0.185 & 0.808 \\
\quad $w^{+}=0.2$  & 0.337 & 0.983 & 0.186 & 0.754 \\
\hdashline
\quad Range (max$-$min) & 0.074 & 0.080 & 0.009 & 0.054 \\
\midrule
\multicolumn{5}{l}{EAPO} \\
\quad $w_{0}=0.05$ & 0.321 & 0.943 & 0.195 & 0.806 \\
\quad $w_{0}=0.1$  & 0.319 & 0.961 & 0.184 & 0.800 \\
\quad $w_{0}=0.2$  & 0.330 & 0.973 & 0.187 & 0.798 \\
\hdashline
\quad Range (max$-$min) & 0.011 & 0.030 & 0.011 & 0.008 \\
\bottomrule
\end{tabular}
}
\caption{Sensitivity of W-REINFORCE and EAPO to the positive-sample weight. Range denotes the gap between the maximum and minimum values across the three settings.}
\label{tab:sensitivity}
\end{table}

\section{Different Fixed Weighting Configurations}
\label{appendix_weight}

Table \ref{tab:fixed-weight-ablation} and Table \ref{tab:fixed-weight-ablation-cmd} respectively present the performance comparison of different methods under various configurations of positive and negative sample weights.

The substantial performance variance observed across different weighting configurations of the same method indicates that fixed-weight schemes struggle to consistently attain optimal performance, as their effectiveness is heavily contingent upon exhaustive hyperparameter search. In contrast, EAPO circumvents explicit weight tuning through an adaptive weighting mechanism, consistently approaching the performance upper bound attained by the best fixed-weight configuration across diverse benchmarks. By adaptively balancing exploration and exploitation, EAPO demonstrates strong practical utility and generalization capability as a tuning-free optimization framework.

We further conduct a comparative analysis of the final performance of EAPO and W-REINFORCE under varying initial positive-sample weights, with the experimental results reported in Table~\ref{tab:sensitivity}.

W-REINFORCE adopts a fixed positive-sample weight $w^{+}$ that remains unchanged throughout training; consequently, its final performance is highly sensitive to this hyperparameter: varying $w^{+}$ alone lifts Rouge-L on RUJA from $0.263$ to $0.337$ and Reranker on RUJA from $0.903$ to $0.983$, indicating that an improper choice of $w^{+}$ can lead to a substantial degradation in performance.

In contrast, EAPO introduces an entropy-feedback adaptive mechanism: the positive-sample weight is decreased during the entropy-decreasing phase to preserve exploration, and increased during the entropy-increasing phase to reinforce stability. Regardless of the value of the initial weight $w_{0}$, the policy entropy is regulated toward a similar trajectory, so that the effective weight gradually converges to a self-adaptive ``operating regime'' determined by the training dynamics rather than by the hyperparameter itself. As a result, the choice of $w_{0}$ has only a marginal influence on the final performance, with the range across all settings remaining within $0.030$.

\section{Cross-Model Generalization}
\label{appendix_generalization}

Table \ref{tab:backbone-comparison} further demonstrates the cross-model generalization capability of EAPO. On Qwen2.5-7B-Instruct, EAPO achieves the best performance across four key metrics, Rouge-L, Reranker, RL@8, and Avg (0.675), consistently outperforming GRPO, PSR, NSR, and W-REINFORCE. EAPO is further amplified on the stronger backbone, indicating that the adaptive weighting mechanism scales favorably with model capacity without requiring any hyperparameter retuning. These results substantiate the robustness and practical utility of EAPO.

\begin{table}[t]
\centering
\small
\resizebox{0.5\textwidth}{!}{%
\setlength{\tabcolsep}{6pt}
\begin{tabular}{lccccc}
\toprule
Methods & Rouge-L & Reranker & RL@8 & RR@8 & Avg \\
\midrule
GRPO        & 0.323 & 0.968 & 0.346 & 0.990 & 0.657 \\
PSR         & 0.287 & 0.944 & 0.294 & 0.970 & 0.624 \\
NSR         & 0.246 & 0.858 & 0.359 & \textbf{0.994} & 0.614 \\
W-REINFORCE & 0.336 & 0.973 & 0.371 & 0.992 & 0.668 \\
\textbf{EAPO} & \textbf{0.341} & \textbf{0.977} & \textbf{0.388} & 0.994 & \textbf{0.675} \\
\bottomrule
\end{tabular}
}
\caption{Performance comparison on Qwen2.5-7B-Instruct. Benchmark: RJUA.}
\label{tab:backbone-comparison}
\end{table}

\end{document}